\documentclass[letterpaper, 10 pt, conference]{ieeeconf}  

\IEEEoverridecommandlockouts                              

\usepackage{graphics} 
\usepackage{graphicx}
\usepackage{amsmath} 
\usepackage{amssymb}  
\usepackage{color}
\usepackage{cite}

\usepackage{multirow}
\usepackage[table,xcdraw]{xcolor}

\title{\LARGE \bf
Cross-modal Place Recognition in Image Databases\\using Event-based Sensors
}

\author{Xiang Ji$^{1\star}$, Jiaxin Wei$^{2\star}$, Yifu Wang$^{2}$, Huiliang Shang$^{1\dag}$ and Laurent Kneip$^{2\dag}$
\thanks{$\star$Authors contributed equally to this work\quad \quad $\dag$ corresponding author}
\thanks{$^{1}$School of Information Science and Technology, Academy for Engineering and Technology, Fudan University {\tt\small \{xji20, shanghl\}@fudan.edu.cn}}%
\thanks{$^{2}$Mobile Perception Lab, School of Information Science and Technology, ShanghaiTech University {\tt\small \{weijx,lkneip\}@shanghaitech.edu.cn, 1fwang927@gmail.com}}%
}

\begin{document}

\maketitle
\thispagestyle{empty}
\pagestyle{empty}

\begin{abstract}
Visual place recognition is an important problem towards global localization in many robotics tasks. One of the biggest challenges is that it may suffer from illumination or appearance changes in surrounding environments. Event cameras are interesting alternatives to frame-based sensors as their high dynamic range enables robust perception in difficult illumination conditions. However, current event-based place recognition methods only rely on event information, which restricts downstream applications of VPR. In this paper, we present the first cross-modal visual place recognition framework that is capable of retrieving regular images from a database given an event query. Our method demonstrates promising results with respect to the state-of-the-art frame-based and event-based methods on the Brisbane-Event-VPR dataset under different scenarios. We also verify the effectiveness of the combination of retrieval and classification, which can boost performance by a large margin.
\end{abstract}

\section{INTRODUCTION}
Visual place recognition (VPR) with conventional cameras has long been studied in both computer vision and robotics communities. It is critical for both 6DoF localization~\cite{PTAM} and loop closure detection in SLAM systems~\cite{orb-slam}\cite{lsd-slam}, and thus helps to enable immersive experiences in AR/VR and autonomous driving in GPS-denied environments, respectively.

VPR is usually formulated as an image retrieval problem~\cite{netvlad}, a task that searches in an image database to retrieve the most similar ones to a query image. Previously seen places are stored as images and each one of them is assigned a location identifier, which can be used to localize the query place later on. However, most frame-based methods struggle in real-world environments where appearance changes are very common over any period of time (e.g. moving objects, day-night differences, weather, and seasonal variations~\cite{vpr-survey}).

The emergence of event cameras points out a promising avenue towards solving this problem. Unlike frame-based sensors producing intensity images at a fixed rate, an event camera is a novel bio-inspired vision sensor that asynchronously detects intensity changes at each pixel. Each time the intensity change surpasses a defined level, an event is fired. An event is then represented as a tuple of 2D pixel coordinates, a timestamp, and a polarity indicating the direction of the intensity change. The invariance to scene illumination~\cite{event-survey} makes them particularly suitable for the VPR task, and several event-based VPR methods~\cite{event-ensemble,event-vpr,eventvlad,HowManyEvents} have been recently proposed to deal with challenging illumination changes.

While using events leads to impressive resilience with respect to appearance changes, it is worth noting that the reference database adopted by those event-based VPR methods is only formed by event data captured under highly similar dynamics. If a place is revisited under different dynamics, fired events will change drastically, thus it is not clear whether such methods would still succeed. Moreover, purely event-based methods are of low practical value in terms of downstream applications compared with an image database. For example, large-scale Structure-from-Motion (SfM)~\cite{sfm} is often solved upon image databases thereby enabling further 6DoF pose estimation~\cite{Liu2017EfficientG2}.

In this work, we bridge the gap between events and real-world images and propose the first cross-modal place recognition pipeline that robustly localizes event queries in a large database of static imagery. Our end-to-end network consists of a global feature extraction backbone to capture image-wide information, a retrieval branch to obtain initial results, and a cross-modal fusion layer followed by a simple MLP-based classifier to further predict the similarity score between two different input modalities (i.e. events and frames). Through extensive experiments on the Brisbane-Event-VPR dataset~\cite{event-ensemble}, we demonstrate competitive performance with respect to the state-of-the-art frame-based and event-based VPR approaches. Despite the increased challenge stemming from the cross-model matching, our method still takes advantage of the illumination-resilient property of events and manages to beat frame-based methods in some difficult illumination conditions. We furthermore demonstrate the benefits of the additional cross-modal fusion and classification over the naive implementation of simply applying a retrieval network for cross-modal VPR.


The main contributions of this paper are as follows:
\begin{enumerate}
    \item We present the first end-to-end pipeline for event-based cross-modal place recognition in image databases. Events possess the advantageous property of being less affected by varying illumination conditions while performing in the traditional image databases easily enables downstream vision tasks such as 6DoF pose estimation.
    \item We combine the weakly supervised representation learning and the fully supervised matching function learning with a cross-modal fusion layer, which is shown to be effective at achieving better results for the VPR task.
    \item We conduct comprehensive experiments on the Brisbane-Event-VPR dataset~\cite{event-ensemble} and show that our method successfully handles a variety of different scenarios and demonstrates competitive performance with respect to frame-based VPR methods.
\end{enumerate}

\section{RELATED WORK}

We start by reviewing some classical frame-based VPR approaches and recently proposed event-based solutions. Then, we will give a brief introduction to bilinear pooling used for cross-modal fusion.

\subsection{Frame-based Visual Place Recognition}

Place recognition is a fundamental problem in robotics and computer vision applications. The most seminal contribution to visual place recognition goes back to \textit{Video Google}~\cite{sivic03}. This work introduces the concept of histograms over visual words, a global image descriptor for retrieving similar images. The use of the so-called bags of keypoints or words has also been introduced by Csurka et al.~\cite{csurka04} and extended by~\cite{nister06, galvez12}. J\'egou et al.~\cite{jegou10} and Arandjelovic et al.~\cite{arandjelovic13} have furthermore proposed aggregation methods to compress local descriptors into compact whole image representations.

The community has continued the development of learning-based alternatives to global image representations for place recognition. Examples are given by the works of Arandjelovic et al.~\cite{netvlad}, Miech et al.~\cite{miech17}, and Ong et al.~\cite{ong18}, who introduce deep networks named \textit{NetVLAD}, \textit{NetVLAD-CG} (NetVLAD with context gating) or \textit{NetBOW}, respectively. The most recent efforts in using learning-based representations for place recognition focus on coarse-to-fine approaches~\cite{sarlin2019coarse,sarlin20} or multi-scale approaches in which local and global features are fused for place recognition-related tasks~\cite{cao20,hausler22netvlad}. More detailed advantages and challenges of VPR are well explained by the original work of Lowry et al.~\cite{vpr-survey} as well as the recent survey by Garg et al.~\cite{Garg2021survey}. While a number of successful works have already been presented, it remains challenging for frame-based solutions to deal with varying illumination conditions.

\subsection{Event-based Visual Place Recognition}

Owing to the superior abilities of event-based sensors, the community has recently stepped up efforts towards visual localization and SLAM with dynamic vision sensors~\cite{milford15,rebecq2016evo,kim2016real,zihao2017event,gallego2017event,mueggler2018continuous,zhou2021event,zuo2022devo,wang2022visual,liu2022icra}. In order to gain invariance with respect to the generally unknown camera dynamics, Fischer and Milford~\cite{event-ensemble} propose to use ensembles of temporal windows of events for place recognition. In their further work~\cite{HowManyEvents}, they analyze the number of events needed for robust place recognition which thereby implicitly adjusts the length of the temporal window according to the captured amount of information.

Recently, the community has proposed an increasing number of network-based solutions for event-based VPR. The most straightforward solution consists of reconstructing images from events \cite{Rebecq19,Mostafavi21} and appending traditional, image-based VPR. However, this approach is prone to fail as it does not describe the input in an appearance-invariant domain. As an alternative, Lee and Kim~\cite{eventvlad} propose EventVLAD, a network to reconstruct gradient maps that are then fed to a NetVLAD architecture, thereby achieving invariance versus day-to-night illumination changes. However, the method uses events that are simulated using Carla, and no direct cross-modal registration ability between events and a database of real-world images was demonstrated. Hussaini et al.~\cite{hussaini22} propose a method based on Spiking Neural Networks (SNNs), which includes a weighting scheme that down-weights the influence of ambiguous neurons responding to multiple different reference places. While interesting, SNN approaches are currently unable to match the maturity of DNNs. Kong et al.~\cite{event-vpr} use EST voxel grid-based event representation~\cite{EST} appended by ResNet and VLAD layers to generate compact descriptors for short event sequences. The network is trained end-to-end based on the triplet ranking loss. However, it again does not perform direct cross-modal registration but matches short sequences of events that are collected under highly similar dynamics. Registration with respect to a database of static imagery is not possible. Hou et al.~\cite{hou2022fe} propose to fuse frames and events to take advantage of two modalities. Their approach consists of attention-based, parallel multi-scale feature extraction from events and images, and subsequent multi-scale fusion and descriptor aggregation. The method involves a heavy network architecture that is unable to run in real time. A similarly heavy network-based approach was presented by Huang et al.~\cite{Huang2022VEF}, who introduce a sequential application of a cross-modality attention module, a self-attention module, and a concluding pooling layer. Their method represents the current state-of-the-art and outperforms both frame and event-based alternatives.


To the best of our knowledge, we present the first cross-modal VPR framework that addresses the challenging problem of directly querying events on top of a database of static images.


\begin{figure*}[!t]
\centering
\includegraphics[width=\linewidth]{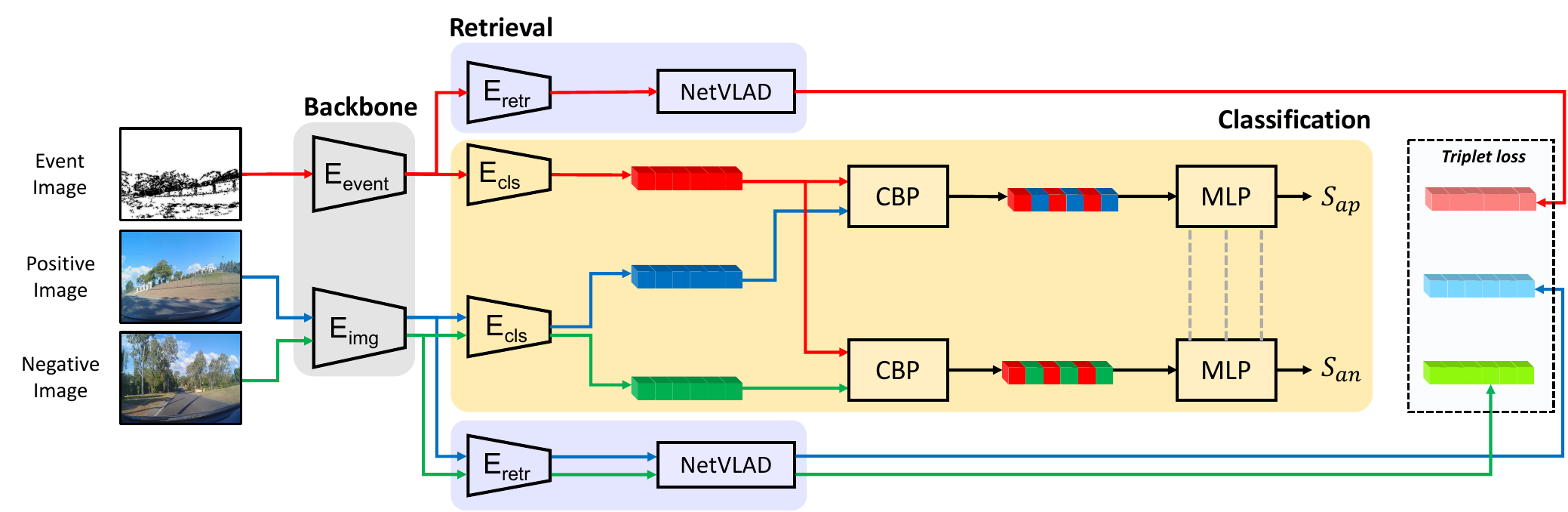}
\caption{The training pipeline of our event-RGB cross-modal VPR. Given our goal of retrieving the most similar RGB image for an event query, we construct a triplet containing an anchor event frame $I_a$, a positive image $I_p$, and a negative image $I_n$ to feed into the network. Our method can be divided into three parts, namely, backbone, retrieval, and classification. During inference, we first perform image retrieval to obtain initial results and then re-rank them using similarity classification.}
\label{Fig:pipeline}
\end{figure*}

\subsection{Bilinear Pooling}

The simplest approaches for merging modalities are vector concatenation and element-wise sum. To improve the fusion quality, Tenenbaum et al. propose a pooling method, called bilinear pooling~\cite{BP}, which calculates the outer product of two vectors such that elements in each vector can have a multiplicative interaction with another vector. Lin et al.~\cite{BilinearCNN} apply bilinear pooling to the fine-grained visual recognition task in order to fuse the feature vectors from two CNNs, achieving large improvements on this task. However, the high dimensionality (i.e. $n^2$) of bilinear pooling limits its use in memory and time-intensive applications. Therefore, Gao et al.~\cite{CompactBP} propose two compact bilinear features based on polynomial kernel analysis, which preserve the same discriminative power but with significantly reduced dimensions. Fukui et al.~\cite{MultimodalCB} then extend it to the question-answering task for vision and text modalities. In this paper, we also use compact bilinear pooling to combine the feature vectors extracted from images and event frames. This fusion step followed by a binary classification can boost retrieval performance by a large margin.

\section{METHOD}

We start by providing an overview of our method before going into the details of the event frame preparation and our network-based representation learning. We conclude with details on our matching function learning and the implementation of the framework.

\subsection{Overview}
The pipeline of our event-RGB cross-modal VPR framework is shown in Figure \ref{Fig:pipeline}. It can be divided into three parts, namely backbone, representation learning (i.e. retrieval), and matching function learning (i.e. classification). The backbone ($E_{img}$ or $E_{event}$) is shared by two following small sub-networks ($E_{retr}$ and $E_{cls}$) to extract global features for retrieval and classification, respectively. \textit{Representation learning} is widely used in both frame-based~\cite{netvlad} and event-based~\cite{event-vpr,eventvlad} VPR methods. It is simple and effective but there is no interaction between the different modalities. To overcome this shortage, we further add \textit{matching function learning}, which comprises a cross-modal fusion layer (i.e. CBP) and a multilayer perceptron (MLP) to improve on the initial retrieval results. While similar in spirit to a recommender system~\cite{deepcf}, our framework represents the first VPR work adopting matching function learning to improve over simple representation learning. Next, we will describe the two key components in detail. 

\subsection{Event Frame Generation}
There are several ways to convert the raw asynchronous event stream into a more suitable representation for later processing, such as event frames~\cite{brebion2021distance_surface}, voxel grid~\cite{VoxelGrid} and event spike tensor~\cite{EST}. Given that we aim at cross-modal retrieval in an image database, we choose the most related representation of event frames~\cite{brebion2021distance_surface} for our events. Specifically, events within a fixed period of time are accumulated into a frame to form an event image, where the intensity value is represented by the spatial distance between the pixel and its nearest event. Such representation reduces the motion dependency of the event and preserves the dense texture information of the scenario. Furthermore, denoising and filling techniques are applied to the event image to enhance the quality and stability of the representation.


\subsection{Network Backbone}
Our network is built upon VGG16~\cite{simonyan2014vgg16}, a deep convolutional neural network architecture for large-scale image classification tasks. Here we use it as our backbone to extract high-level information from both event frames and RGB images. The output feature maps are then fed into two small networks to generate global feature vectors for subsequent retrieval and classification, respectively. Let $f_\theta$ denote the backbone. We have
\begin{equation*}
    F_i = f_\theta(I_i),
\end{equation*}
where $I_i$ is either an event frame or an input RGB image.

\subsection{Representation Learning}
We map the input images and events into the representation space, where they can be directly compared using distance metrics.

\textbf{Global Feature Extraction for Retrieval.}
We put a NetVLAD layer~\cite{netvlad} after a small sub-network $E_{retr}$ such that it aggregates mid-level feature maps into a powerful feature vector encoding the global image-wide information. This sub-network only consists of three convolutional layers, and we denote it $f_\phi$. The output for retrieval is then given by
\begin{equation*}
    F_i^{retr} = \mathrm{NetVLAD}(f_\phi(F_i)) \in \mathbb{R}^{(K\times D)\times 1},
\end{equation*}
where $K$ is the VLAD parameter indicating the number of cluster centers, and $D$ is the dimension of local features extracted by CNNs.

\textbf{Triplet Loss.}
Towards the goal of matching query events in image databases, a training triplet $\{I_a, I_p, I_n\}$ containing an anchor event frame $I_a$, a positive image $I_p$, and a negative image $I_n$, is processed by the retrieval branches in a single forward pass. Thus, the resulting features are as follows
\begin{equation*}
    \{F_a^{retr}, F_p^{retr}, F_n^{retr}\}.
\end{equation*}
The triplet loss is then given by
\begin{equation*}
    L_{triplet} = \mathrm{max}(d(F_a^{retr}, F_p^{retr}) - d(F_a^{retr}, F_n^{retr}) + \alpha, 0)
\end{equation*}
where $d(\cdot)$ is the Euclidean distance and $\alpha$ is a margin between positive and negative pairs.

\subsection{Matching Function Learning}

We fuse the triplet features for classification in a pair-wise manner by means of Compact Bilinear Pooling (CBP)~\cite{CompactBP}, and then successively feed the fused features into an MLP-based classifier to predict the similarity score. Thus, a complex matching function is learned by the MLP under strong supervision signals.

\textbf{Intermediate Feature Extraction for Classification.}
The sub-network architecture for classification is quite similar to the one for retrieval. The difference is that $E_{cls}$ has a fully connected layer following the CNNs to produce a lower-dimensional feature vector. Let $f_\omega$ denote the sub-network, then we have
\begin{equation*}
    F_i^{cls} = f_\omega(F_i).
\end{equation*}

\begin{figure}[bp]
\centering
\includegraphics[width=\linewidth]{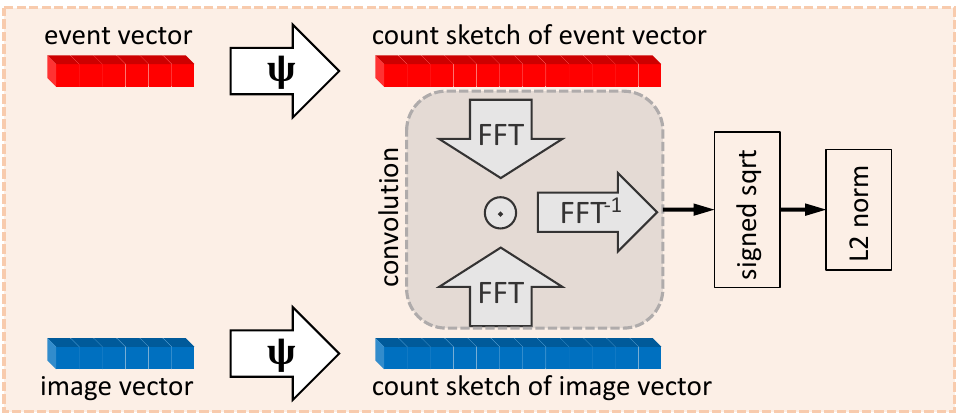}
\caption{Architecture of the CBP module for cross-modal fusion.}
\label{Fig:CBP}
\end{figure}

\textbf{Cross-modal Fusion.}
The CBP algorithm (see Figure \ref{Fig:CBP}) first projects the feature vector $F_i^{cls} \in \mathbb{R}^{n}$ to $\hat{F}_i^{cls} \in \mathbb{R}^{m}$ using the Count Sketch projection function $\Psi$ \cite{CountSketch}, i.e. $\hat{F}_i^{cls} = \Psi(F_i^{cls})$. Then, the fusion of two feature vectors can be formulated as 
\begin{equation*}
    \hat{F}_{ij}^{cls} = \mathrm{FFT}^{-1}(\mathrm{FFT}(\hat{F}_i^{cls}) \odot \mathrm{FFT}(\hat{F}_j^{cls})),
\end{equation*}
where $\odot$ refers to element-wise product, and $\mathrm{FFT}$ and $\mathrm{FFT}^{-1}$ represent the Fast Fourier Transform and its inverse, respectively. As for the training triplet, we construct the anchor-positive pair $\{F_a^{cls}, F_p^{cls}\}$ and the anchor-negative pair $\{F_a^{cls}, F_n^{cls}\}$, which are passed through the CBP module to get the corresponding fused features $\hat{F}_{ap}^{cls}$ and $\hat{F}_{an}^{cls}$.

\textbf{Similarity Classification.}
To determine whether or not the two inputs are similar to each other, we use a 3-layer MLP, denoted as $g_{\tau}$, to approximate the complex matching function. Hence, 
\begin{equation*}
    S_{ap} = g_{\tau}(\hat{F}_{ap}^{cls}) ~\mathrm{and}~ S_{an} = g_{\tau}(\hat{F}_{an}^{cls})
\end{equation*}
are the two predicted matching scores. The parameters of MLP are optimized by minimizing a binary cross-entropy loss. Note that it is easy to get the ground truth labels for classification since we know the relations among the training triplet. Therefore, the loss function for similarity classification is 
\begin{equation*}
    L_{cls} = \mathrm{BCE}(S_{ap}, 1) + \mathrm{BCE}(S_{an}, 0).
\end{equation*}

\subsection{Implementation Details}

\textbf{Loss Function.}
The overall loss function for end-to-end training is the sum of the triplet and the classification loss
\begin{equation*}
    L = L_{triplet} + L_{cls}.
\end{equation*}

\textbf{Network Training.}
For event frame generation, we use a temporal window of size $\Delta T = 25ms$ and follow the steps described in \cite{brebion2021distance_surface}. The image resolution of DAVIS346 is $346 \times 260$, but the resolution of the RGB images from the consumer camera is $640 \times 480$. Therefore, we resize all the images with higher resolution to be consistent with event frames and DAVIS RGB images.

The number of cluster centers $K$ in the NetVLAD layer is set to 64 and the dimension of local features $D$ is 512, resulting in a $(64 \times 512)$-dim feature vector for retrieval. The intermediate feature for classification is 4096-dim. It will be fused with another feature vector by the CBP module to obtain a 1024-dim feature for final similarity classification. The margin $\alpha$ in the triplet loss is set to 0.1, and we use a SGD optimizer with a learning rate of 0.1 to train the whole network in an end-to-end manner.

\textbf{Inference.}
During inference, the workflow is slightly different from training. The image database is pre-processed offline such that each RGB image has two feature vectors associated with it, one for retrieval and the other one for classification. Given an event query (i.e. event frame), we first perform image retrieval in the database to obtain the initial result which is sorted in descending order according to the feature distances. After narrowing down the search scope, we then re-rank the initial result by computing the similarity scores between the query and each image candidate. Therefore, the final result is several retrieval candidates sorted in descending order according to their similarity scores.

\section{EXPERIMENTS}
In this section, we first conduct experiments on an event-based VPR dataset and compare our method with frame-based and event-based approaches quantitatively to demonstrate the performance under different scenarios. Next, we verify the effectiveness of cross-modal fusion and classification by comparing it with the naive implementation of simple retrieval. Finally, we discuss the limitations of our method and point out future directions to further improve it.

\subsection{Dataset}

\begin{figure}[b!]
\centering
\includegraphics[width=\linewidth]{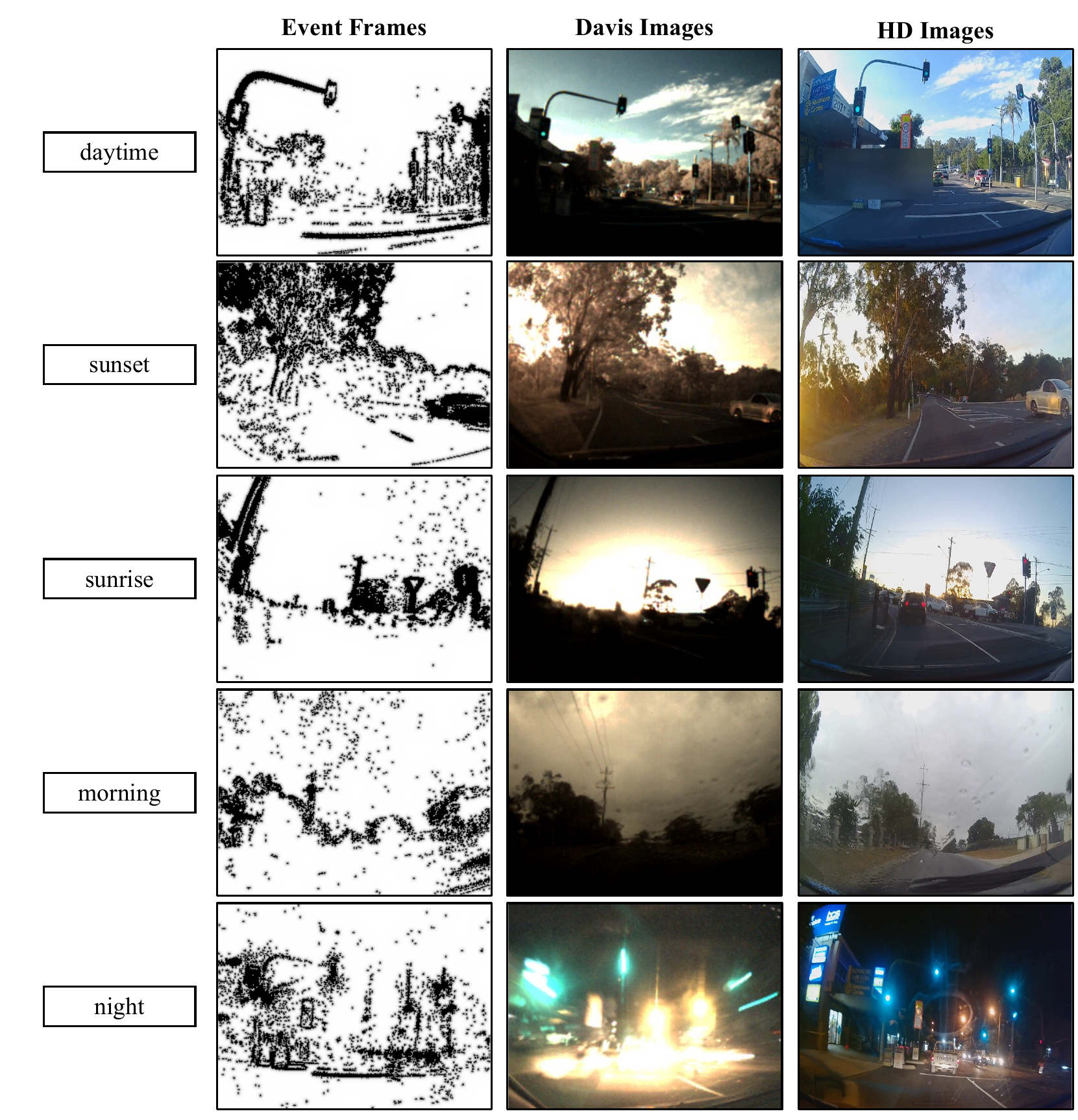}
\caption{Visualizations of samples from the Brisbane-Event-VPR Dataset.}
\label{Fig:data_sample}
\end{figure}

The Brisbane-Event-VPR dataset~\cite{event-ensemble} provides a benchmark for evaluating the performance of event-based VPR algorithms. It is captured in an outdoor environment and covers a distance of approximately 8 km along a route captured under 6 different scenarios with changing illumination conditions (i.e. daytime, sunset, sunrise, morning, and night). We divide each traverse into 3 splits that are geographically non-overlapping and use them for training, validation, and testing, respectively. The specific configuration is listed in Table~\ref{Tab:dataset}. Except for the event data, this dataset also contains RGB images collected by an event camera (DAVIS346) and a consumer camera. Therefore, each data sample comprises a converted event frame, a DAVIS RGB image, a high-quality RGB image, and a GPS coordinate. Some examples are shown in Figure~\ref{Fig:data_sample}.

\begin{table}[htbp]
\begin{tabular}{ccccccc}
\hline
\multicolumn{7}{c}{Brisbane-Event-VPR Dataset}                                                                                  \\ \hline
\multicolumn{1}{c|}{} & daytime & sunset & sunrise & morning & \multicolumn{1}{c|}{night} & Total \\ \hline
\multicolumn{1}{c|}{Train}                          & 5989    & 4826   & 5569    & 5506    & \multicolumn{1}{c|}{5579}  & 27469 \\
\multicolumn{1}{c|}{Val}                            & 499     & 520    & 550     & 530     & \multicolumn{1}{c|}{550}   & 2649  \\
\multicolumn{1}{c|}{Test}                           & 501     & 531    & 551     & 541     & \multicolumn{1}{c|}{561}   & 2685  \\ \hline
\multicolumn{1}{c|}{Total}                          & 6989    & 5877   & 6670    & 6577    & \multicolumn{1}{c|}{6690}  & 32803 \\ \hline
\end{tabular}
\caption{Taxonomy of the Brisbane-Event-VPR Dataset~\cite{event-ensemble}.}
\label{Tab:dataset}
\end{table}

\subsection{Comparisons}
Given that our cross-modal method naturally includes a frame branch and an event branch for retrieval, these two branches can work independently and serve as our frame-based and event-based references, respectively. Although these two branches both possess a network architecture similar to NetVLAD~\cite{netvlad}, they are trained with different data formats (i.e. event frames and RGB images). Here we carry out experiments on both high-quality RGB images and DAVIS RGB images in order to fully demonstrate the performance of the different methods. The database for retrieval is built using the test set from the daytime, and we use test samples from all 6 scenarios as queries to search the database.

\begin{table}[b!]
\centering
\begin{tabular}{c|cccccc}
\hline
                                                                          & \multicolumn{1}{c|}{}                    & \multicolumn{5}{c}{Scenarios}                                                                                                                            \\ \cline{3-7} 
\multirow{-2}{*}{Methods}                                                 & \multicolumn{1}{c|}{\multirow{-2}{*}{N}} & daytime                      & sunset                       & sunrise                      & morning                      & night                        \\ \hline
                                                                          & \multicolumn{1}{c|}{1}                   & \cellcolor[HTML]{FFCCC9}1.00 & \cellcolor[HTML]{FFCCC9}0.52 & \cellcolor[HTML]{9AFF99}0.55 & \cellcolor[HTML]{9AFF99}0.42 & \cellcolor[HTML]{9AFF99}0.36 \\
                                                                          & \multicolumn{1}{c|}{5}                   & \cellcolor[HTML]{FFCCC9}1.00 & \cellcolor[HTML]{FFCCC9}0.69 & \cellcolor[HTML]{96FFFB}0.72 & \cellcolor[HTML]{9AFF99}0.53 & \cellcolor[HTML]{9AFF99}0.52 \\
                                                                          & \multicolumn{1}{c|}{10}                  & \cellcolor[HTML]{FFCCC9}1.00 & \cellcolor[HTML]{FFCCC9}0.81 & \cellcolor[HTML]{96FFFB}0.77 & \cellcolor[HTML]{9AFF99}0.62 & \cellcolor[HTML]{9AFF99}0.59 \\
                                                                          & \multicolumn{1}{c|}{15}                  & \cellcolor[HTML]{FFCCC9}1.00 & \cellcolor[HTML]{FFCCC9}0.92 & \cellcolor[HTML]{96FFFB}0.79 & \cellcolor[HTML]{9AFF99}0.69 & \cellcolor[HTML]{9AFF99}0.63 \\
                                                                          & \multicolumn{1}{c|}{20}                  & \cellcolor[HTML]{FFCCC9}1.00 & \cellcolor[HTML]{FFCCC9}0.95 & \cellcolor[HTML]{CBCEFB}0.83 & \cellcolor[HTML]{9AFF99}0.75 & \cellcolor[HTML]{CBCEFB}0.65 \\
\multirow{-6}{*}{\begin{tabular}[c]{@{}c@{}}RGB \\ (camera)\end{tabular}} & \multicolumn{1}{c|}{30}                  & \cellcolor[HTML]{FFCCC9}1.00 & \cellcolor[HTML]{FFCCC9}0.97 & \cellcolor[HTML]{CBCEFB}0.87 & \cellcolor[HTML]{9AFF99}0.85 & \cellcolor[HTML]{CBCEFB}0.69 \\ \hline
                                                                          & \multicolumn{1}{c|}{1}                   & \cellcolor[HTML]{FFCCC9}1.00 & \cellcolor[HTML]{9AFF99}0.49 & \cellcolor[HTML]{FFCCC9}0.65 & \cellcolor[HTML]{FFCCC9}0.43 & \cellcolor[HTML]{FFCCC9}0.38 \\
                                                                          & \multicolumn{1}{c|}{5}                   & \cellcolor[HTML]{FFCCC9}1.00 & \cellcolor[HTML]{96FFFB}0.59 & \cellcolor[HTML]{9AFF99}0.80 & \cellcolor[HTML]{96FFFB}0.49 & \cellcolor[HTML]{96FFFB}0.51 \\
                                                                          & \multicolumn{1}{c|}{10}                  & \cellcolor[HTML]{FFCCC9}1.00 & \cellcolor[HTML]{96FFFB}0.66 & \cellcolor[HTML]{9AFF99}0.86 & \cellcolor[HTML]{CBCEFB}0.55 & \cellcolor[HTML]{96FFFB}0.58 \\
                                                                          & \multicolumn{1}{c|}{15}                  & \cellcolor[HTML]{FFCCC9}1.00 & \cellcolor[HTML]{CBCEFB}0.68 & \cellcolor[HTML]{9AFF99}0.88 & \cellcolor[HTML]{CBCEFB}0.60 & \cellcolor[HTML]{9AFF99}0.63 \\
                                                                          & \multicolumn{1}{c|}{20}                  & \cellcolor[HTML]{FFCCC9}1.00 & \cellcolor[HTML]{CBCEFB}0.72 & \cellcolor[HTML]{9AFF99}0.92 & \cellcolor[HTML]{CBCEFB}0.63 & \cellcolor[HTML]{96FFFB}0.68 \\
\multirow{-6}{*}{\begin{tabular}[c]{@{}c@{}}RGB \\ (DAVIS)\end{tabular}}  & \multicolumn{1}{c|}{30}                  & \cellcolor[HTML]{FFCCC9}1.00 & \cellcolor[HTML]{CBCEFB}0.78 & \cellcolor[HTML]{9AFF99}0.94 & \cellcolor[HTML]{CBCEFB}0.70 & \cellcolor[HTML]{96FFFB}0.77 \\ \hline
                                                                          & \multicolumn{1}{c|}{1}                   & \cellcolor[HTML]{FFCCC9}1.00 & \cellcolor[HTML]{96FFFB}0.38 & \cellcolor[HTML]{96FFFB}0.52 & \cellcolor[HTML]{96FFFB}0.35 & \cellcolor[HTML]{96FFFB}0.25 \\
                                                                          & \multicolumn{1}{c|}{5}                   & \cellcolor[HTML]{FFCCC9}1.00 & \cellcolor[HTML]{9AFF99}0.62 & \cellcolor[HTML]{FFCCC9}0.81 & \cellcolor[HTML]{FFCCC9}0.58 & \cellcolor[HTML]{FFCCC9}0.53 \\
                                                                          & \multicolumn{1}{c|}{10}                  & \cellcolor[HTML]{FFCCC9}1.00 & \cellcolor[HTML]{9AFF99}0.73 & \cellcolor[HTML]{FFCCC9}0.89 & \cellcolor[HTML]{FFCCC9}0.72 & \cellcolor[HTML]{FFCCC9}0.67 \\
                                                                          & \multicolumn{1}{c|}{15}                  & \cellcolor[HTML]{FFCCC9}1.00 & \cellcolor[HTML]{9AFF99}0.81 & \cellcolor[HTML]{FFCCC9}0.93 & \cellcolor[HTML]{FFCCC9}0.78 & \cellcolor[HTML]{FFCCC9}0.74 \\
                                                                          & \multicolumn{1}{c|}{20}                  & \cellcolor[HTML]{FFCCC9}1.00 & \cellcolor[HTML]{9AFF99}0.87 & \cellcolor[HTML]{FFCCC9}0.95 & \cellcolor[HTML]{FFCCC9}0.83 & \cellcolor[HTML]{FFCCC9}0.82 \\
\multirow{-6}{*}{\begin{tabular}[c]{@{}c@{}}Event\\ (DAVIS)\end{tabular}} & \multicolumn{1}{c|}{30}                  & \cellcolor[HTML]{FFCCC9}1.00 & \cellcolor[HTML]{9AFF99}0.92 & \cellcolor[HTML]{FFCCC9}0.97 & \cellcolor[HTML]{FFCCC9}0.88 & \cellcolor[HTML]{FFCCC9}0.92 \\ \hline
                                                                          & \multicolumn{1}{c|}{1}                   & \cellcolor[HTML]{9AFF99}0.36 & \cellcolor[HTML]{CBCEFB}0.21 & \cellcolor[HTML]{CBCEFB}0.21 & \cellcolor[HTML]{CBCEFB}0.17 & \cellcolor[HTML]{CBCEFB}0.22 \\
                                                                          & \multicolumn{1}{c|}{5}                   & \cellcolor[HTML]{9AFF99}0.67 & \cellcolor[HTML]{CBCEFB}0.50 & \cellcolor[HTML]{CBCEFB}0.43 & \cellcolor[HTML]{CBCEFB}0.46 & \cellcolor[HTML]{CBCEFB}0.40 \\
                                                                          & \multicolumn{1}{c|}{10}                  & \cellcolor[HTML]{9AFF99}0.83 & \cellcolor[HTML]{CBCEFB}0.62 & \cellcolor[HTML]{CBCEFB}0.60 & \cellcolor[HTML]{96FFFB}0.60 & \cellcolor[HTML]{CBCEFB}0.52 \\
                                                                          & \multicolumn{1}{c|}{15}                  & \cellcolor[HTML]{9AFF99}0.90 & \cellcolor[HTML]{96FFFB}0.75 & \cellcolor[HTML]{CBCEFB}0.75 & \cellcolor[HTML]{96FFFB}0.66 & \cellcolor[HTML]{96FFFB}0.61 \\
                                                                          & \multicolumn{1}{c|}{20}                  & \cellcolor[HTML]{9AFF99}0.92 & \cellcolor[HTML]{96FFFB}0.86 & \cellcolor[HTML]{96FFFB}0.85 & \cellcolor[HTML]{96FFFB}0.71 & \cellcolor[HTML]{9AFF99}0.69 \\
\multirow{-6}{*}{Ours}                                                    & \multicolumn{1}{c|}{30}                                       & \cellcolor[HTML]{9AFF99}0.94 & \cellcolor[HTML]{96FFFB}0.91 & \cellcolor[HTML]{96FFFB}0.92 & \cellcolor[HTML]{96FFFB}0.76 & \cellcolor[HTML]{9AFF99}0.80 \\ \hline
\end{tabular}
\caption{Comparison Results on Brisbane-Event-VPR dataset.}
\label{Tab:exp_data}
\end{table}

\subsection{Metrics}
Following prior work~\cite{event-vpr}, we use Recall@N to evaluate the performance of different methods. In our experiments, we choose $N=\{1, 5, 10, 15, 20, 30\}$. A match for the given query is considered as positive if the geographic distance is less than 35m.

\begin{figure*}[htbp]
\centering
\includegraphics[width=0.94\linewidth]{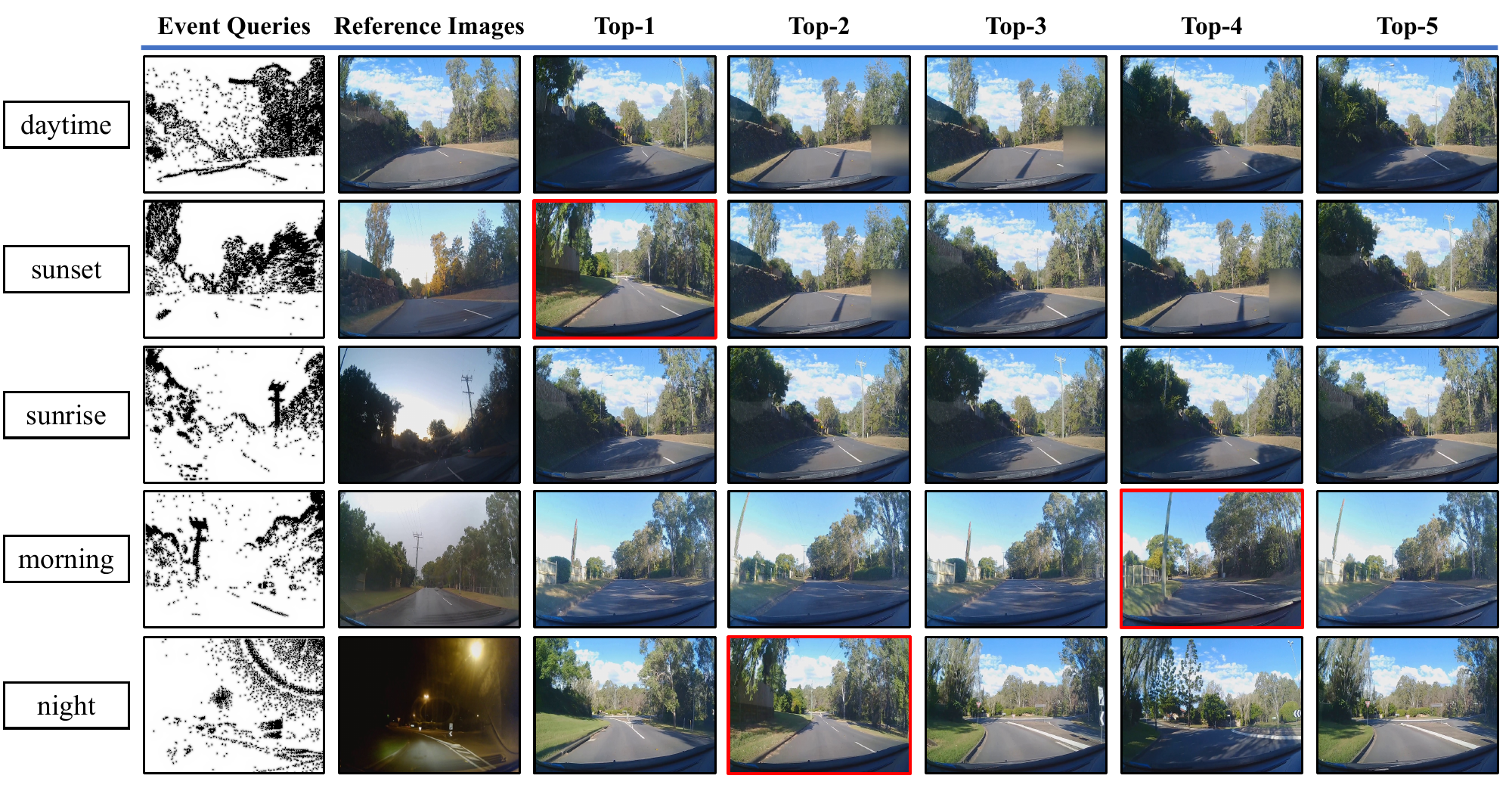}
\caption{Qualitative results of our method on the Brisbane-Event-VPR dataset under different scenarios. High-quality RGB images are also shown for reference. Negative samples are highlighted with a red box.}
\label{Fig:top5_vis}
\end{figure*}

\subsection{Performance in Different Scenarios}
The comparison results on the Brisbane-Event-VPR dataset under different scenarios are shown in Table~\ref{Tab:exp_data}, which is color coded to better visualize the data (red, green, blue, and purple indicate values in descending order). As can be observed, frame-based and event-based methods can achieve a 100\% recall rate in the daytime setting given the query and database samples come from the same source. On contrary, obtaining similarly excellent performance in the cross-modal setting is unrealistic given the discrepancy between event queries and image samples.

From data examples shown in Figure~\ref{Fig:data_sample}, we can discover that in most cases the image quality of a consumer camera is better than a DAVIS camera. Also, the converted event frame is capable of reserving most of the important information in a scene with less interference from illumination conditions. These properties are again proven in Table~\ref{Tab:exp_data}, where the frame-based method using high-quality images outperforms that of using DAVIS images in most of the scenarios, and the event-based method performs better than frame-based methods in poor illumination conditions. Though the cross-modal task is harder than the traditional VPR task, our method still benefits from some of the good properties of events and beats the frame-based methods in some challenging scenarios (see blue areas for our  method). Further qualitative results of our method are indicated in Figure~\ref{Fig:top5_vis}.

\subsection{Ablation Study}
To demonstrate the effectiveness of the proposed cross-modal fusion and classification for re-ranking, we compare our method with a naive alternative, that is, a simple pipeline only applies the retrieval function. The comparison results are shown in Table~\ref{Tab:ablation}. As can be observed, the combination of retrieval and classification can boost performance by a large margin.

\begin{table}[htbp]
\centering
\begin{tabular}{c|c|ccccc}
\hline
                                                &                     & \multicolumn{5}{c}{Scenarios}                                                                                                                            \\ \cline{3-7} 
\multirow{-2}{*}{Methods}                       & \multirow{-2}{*}{N} & daytime                      & sunset                       & sunrise                      & morning                      & night                        \\ \hline
                                                & 1                   & \cellcolor[HTML]{9AFF99}0.11 & \cellcolor[HTML]{9AFF99}0.12 & \cellcolor[HTML]{9AFF99}0.11 & \cellcolor[HTML]{9AFF99}0.12 & \cellcolor[HTML]{9AFF99}0.13 \\
                                                & 5                   & \cellcolor[HTML]{9AFF99}0.38 & \cellcolor[HTML]{9AFF99}0.28 & \cellcolor[HTML]{9AFF99}0.32 & \cellcolor[HTML]{9AFF99}0.30 & \cellcolor[HTML]{9AFF99}0.28 \\
                                                & 10                  & \cellcolor[HTML]{9AFF99}0.70 & \cellcolor[HTML]{9AFF99}0.57 & \cellcolor[HTML]{9AFF99}0.58 & \cellcolor[HTML]{9AFF99}0.53 & \cellcolor[HTML]{9AFF99}0.47 \\
                                                & 15                  & \cellcolor[HTML]{9AFF99}0.84 & \cellcolor[HTML]{FFCCC9}0.75 & \cellcolor[HTML]{FFCCC9}0.75 & \cellcolor[HTML]{FFCCC9}0.68 & \cellcolor[HTML]{9AFF99}0.54 \\
                                                & 20                  & \cellcolor[HTML]{9AFF99}0.89 & \cellcolor[HTML]{9AFF99}0.79 & \cellcolor[HTML]{9AFF99}0.80 & \cellcolor[HTML]{FFCCC9}0.72 & \cellcolor[HTML]{9AFF99}0.58 \\
\multirow{-6}{*}{\begin{tabular}[c]{@{}c@{}}Ours\\ (retrieval)\end{tabular}}              & 30                  & \cellcolor[HTML]{9AFF99}0.92 & \cellcolor[HTML]{9AFF99}0.85 & \cellcolor[HTML]{9AFF99}0.85 & \cellcolor[HTML]{FFCCC9}0.82 & \cellcolor[HTML]{9AFF99}0.65 \\ \hline
                                                & 1                   & \cellcolor[HTML]{FFCCC9}0.36 & \cellcolor[HTML]{FFCCC9}0.21 & \cellcolor[HTML]{FFCCC9}0.21 & \cellcolor[HTML]{FFCCC9}0.17 & \cellcolor[HTML]{FFCCC9}0.22 \\
                                                & 5                   & \cellcolor[HTML]{FFCCC9}0.67 & \cellcolor[HTML]{FFCCC9}0.50 & \cellcolor[HTML]{FFCCC9}0.43 & \cellcolor[HTML]{FFCCC9}0.46 & \cellcolor[HTML]{FFCCC9}0.40 \\
                                                & 10                  & \cellcolor[HTML]{FFCCC9}0.83 & \cellcolor[HTML]{FFCCC9}0.62 & \cellcolor[HTML]{FFCCC9}0.60 & \cellcolor[HTML]{FFCCC9}0.60 & \cellcolor[HTML]{FFCCC9}0.52 \\
                                                & 15                  & \cellcolor[HTML]{FFCCC9}0.90 & \cellcolor[HTML]{FFCCC9}0.75 & \cellcolor[HTML]{FFCCC9}0.75 & \cellcolor[HTML]{9AFF99}0.66 & \cellcolor[HTML]{FFCCC9}0.61 \\
                                                & 20                  & \cellcolor[HTML]{FFCCC9}0.92 & \cellcolor[HTML]{FFCCC9}0.86 & \cellcolor[HTML]{FFCCC9}0.85 & \cellcolor[HTML]{9AFF99}0.71 & \cellcolor[HTML]{FFCCC9}0.69 \\
\multirow{-6}{*}{\begin{tabular}[c]{@{}c@{}}Ours\\ (hybrid)\end{tabular}} & 30                  & \cellcolor[HTML]{FFCCC9}0.94 & \cellcolor[HTML]{FFCCC9}0.91 & \cellcolor[HTML]{FFCCC9}0.92 & \cellcolor[HTML]{9AFF99}0.76 & \cellcolor[HTML]{FFCCC9}0.80 \\ \hline
\end{tabular}
\caption{Ablation Study of the additional classification module.}
\label{Tab:ablation}
\end{table}

\subsection{Limitations}
Although our hybrid cross-modal pipeline has shown some promising results in the VPR task, there is still a gap in overall performance compared to frame-based and event-based methods. However, it provides a new perspective to utilize both event-based and frame-based sensors. Our purpose is to find an alternative for images in poor illumination conditions (e.g. night), not to totally replace the mature frame-based sensors in all circumstances. Therefore, we will focus on optimizing the network architecture in the future to further improve the retrieval performance and also explore the possibilities of 6DoF pose estimation given an event query.

\section{CONCLUSIONS}

We have proposed the first cross-modal VPR framework that directly retrieves regular images for event queries. Strong benefits over a plain NetVLAD-style retrieval architecture have been obtained by adding a re-ranking module, comprising a compact bilinear pooling layer and a similarity classification network. Though the cross-modal nature of the retrieval task remains a big challenge, our method still achieves some competitive results compared to traditional frame-based alternatives in difficult illumination conditions. We believe that our method provides a new perspective to combine the advantages of two modalities and makes an important contribution towards the practicability of event-based VPR.



\bibliographystyle{IEEEtran}
\bibliography{ref}
\end{document}